\pdfoutput=1

\documentclass[11pt]{article}

\usepackage{emnlp2021}

\usepackage{times}
\usepackage{latexsym}
\usepackage{amsmath}
\usepackage{booktabs}
\usepackage{multirow}
\usepackage{adjustbox}
\usepackage{graphicx}
\usepackage[T1]{fontenc}
\usepackage{subfigure}
\usepackage{tabularx}

\usepackage[utf8]{inputenc}

\usepackage{microtype}

\usepackage{xspace}
\newcommand\BASESIZE{$_{\small \textsc{base}}$\xspace}
\newcommand\LARGESIZE{$_{\small \textsc{large}}$\xspace}
\newlength{\lsuperstar}
\title{Dynamic Knowledge Distillation for Pre-trained Language Models}

\author{Lei Li\textsuperscript{$\dag$}, Yankai Lin\textsuperscript{$\S$}, Shuhuai Ren\textsuperscript{$\dag$}, Peng Li\textsuperscript{$\S$}, Jie Zhou\textsuperscript{$\S$}, Xu Sun\textsuperscript{$\dag$}\\
   \textsuperscript{$\dag$}MOE Key Laboratory of Computational Linguistics, School of EECS, Peking University \\
  \textsuperscript{$\S$}Pattern Recognition Center, WeChat AI, Tencent Inc., China\\
    \texttt{\{lilei, shuhuai\_ren\}@stu.pku.edu.cn}\quad \texttt{xusun@pku.edu.cn} \\
    \texttt{\{yankailin, patrickpli, withtomzhou\}@tecent.com}
  }

\begin{document}
\maketitle
\begin{abstract}
Knowledge distillation~(KD) has been proved effective for compressing large-scale pre-trained language models.
However, existing methods conduct KD statically, e.g., the student model aligns its output distribution to that of a selected teacher model on the pre-defined training dataset.
In this paper, we explore whether a dynamic knowledge distillation that empowers the student to adjust the learning procedure according to its competency, regarding the student performance and learning efficiency.
We explore the dynamical adjustments on three aspects: teacher model adoption, data selection, and KD objective adaptation.
Experimental results show that (1) proper selection of teacher model can boost the performance of student model; (2) conducting KD with 10\% informative instances achieves comparable performance while greatly accelerates the training; (3) the student performance can be boosted by adjusting the supervision contribution of different alignment objective.
We find dynamic knowledge distillation is promising and provide discussions on potential future directions towards more efficient KD methods.\footnote{Our code is available at \url{https://github.com/lancopku/DynamicKD}}
\end{abstract}

\section{Introduction}
Knowledge distillation~(KD)~\citep{Hinton2015Distilling} aims to transfer the knowledge from a large teacher model to a small student model. It has been widely used~\citep{Sanh2019DistilBERT,Jiao2019TinyBERT,Sun2019PatientKD} to compress large-scale pre-trained language models~(PLMs) like BERT~\citep{devlin2019bert} and RoBERTa~\citep{Liu2019RoBERTa} in recent years. By knowledge distillation, we can obtain a much smaller model with comparable performance, while greatly reduce the memory usage and accelerate the model inference.


Although simple and effective, existing methods usually conduct the KD learning procedure statically, e.g., the student model aligns its output probability distribution to that of a selected teacher model on the entire pre-defined corpus. 
In other words, the following three aspects of KD are specified in advance and remain unchanged during the learning procedure: (1) the teacher model to learn from (learning target); (2) the data used to query the teacher (learning material); (3) the objective functions and the corresponding weights (learning method). However, as the student competency evolves during the training stage, it is particularly unreasonable to pre-define these learning settings and keep them unchanged. 
Conducting KD statically may lead to (1) unqualified learning from a too large teacher, (2) repetitive learning on instances that the student has mastered, and (3) sub-optimal learning on alignments that are unnecessary. This motivates us to explore an interesting problem: whether a dynamic KD framework considering the student competency evolution during training can bring benefits, regarding the student performance and learning efficiency?

\begin{figure*}[t!]
    \centering
    \includegraphics[width=0.9\linewidth]{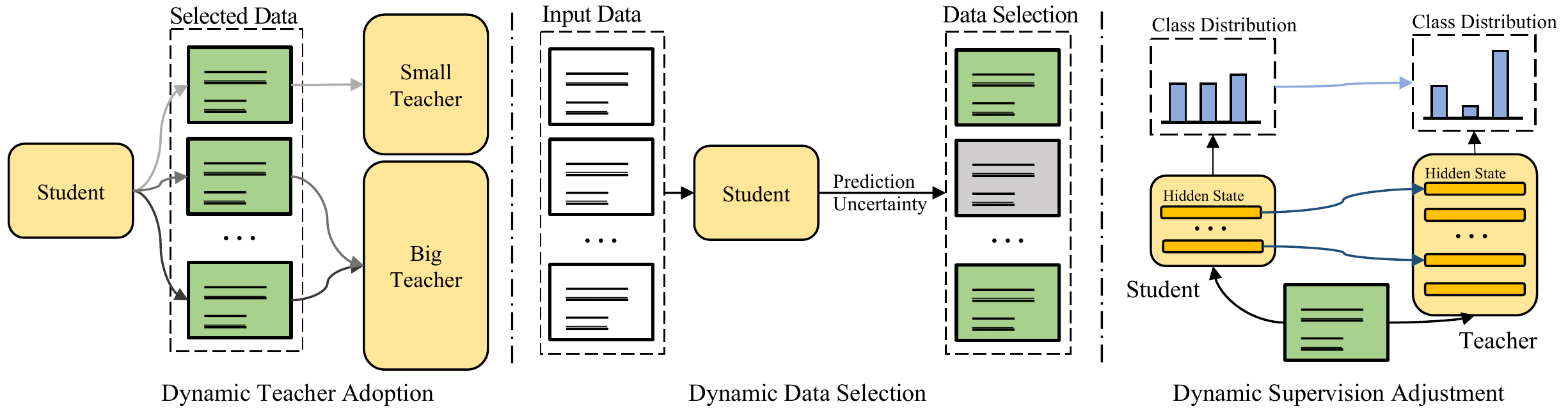}
    \caption{The three aspects of dynamic knowledge distillation explored in this paper. Best viewed in color.}
    \label{fig:dkd}
\end{figure*}




In this paper, we propose a dynamic knowledge distillation~(Dynamic KD) framework, which attempts to 
empower the student to adjust the learning procedure according to its competency.
Specifically, inspired by the success of active learning~\citep{settles2009active}, we take the prediction uncertainty, e.g., the entropy of the predicted classification probability distribution, as a proxy of the student competency. We strive to answer the following research questions: (RQ1) Which teacher is proper to learn as the student evolves? (RQ2) Which data are actually useful for student models in the whole KD stage?  (RQ3) Does the optimal learning objective change in the KD process? In particular, we first explore the impact of the teacher size to dynamic knowledge distillation. Second, we explore whether dynamically choosing instances that the student is uncertain for KD can lead to a better performance and training efficiency trade-off. Third, we explore whether the dynamic adjustment of the supervision from alignment of prediction probability distributions and the alignment of intermediate representations in the whole KD stage can improve the performance.

Our experimental results show that: 
(1) A larger teacher model with more layers may raise a worse student. We show that selecting the proper teacher model according to the competency of the student can improve the performance. 
(2) We can achieve comparable performance using only 10\% informative instances selected according to the student prediction uncertainty. These instances also evolve during the training as the student becomes stronger.
(3) We can boost the student performance by dynamically adjusting the supervision from different alignment objectives of the teacher model. 

Our observations demonstrate the limitations of the current static KD framework. 
The proposed uncertainty-based dynamic KD framework only makes the very first attempt, and we are hoping this paper can motivate more future research towards more efficient and adaptive KD methods.

\section{Background: Knowledge Distillation}
Given a student model $S$ and a teacher model $T$, knowledge distillation aims to train the student model by aligning the outputs of the student model to that of the teacher.
For example, \citet{Hinton2015Distilling} utilize the teacher model outputs as soft targets for the student to learn.
We denote $S(x)$ and $T(x)$ as the output logit vector of the student and the teacher for input $x$, respectively. The KD can be conducted by minimizing the Kullback-Leibler~(KL) divergence distance between the student and teacher prediction:
\begin{equation}
    \mathcal{L}_{KL} =  \text{KL} \left( \sigma\left( S\left(x\right)  / \tau \right) ||  \sigma \left(T \left(x\right) / \tau \right) \right),  
\label{eq:kd_loss}
\end{equation}
where $\sigma(\cdot)$ denotes the softmax function and $\tau$ is a temperature hyper-parameter. The student parameters are updated according to the KD loss and the original classification loss, i.e., the cross-entropy over the ground-truth label $y$:
\begin{align}
    \mathcal{L}_{CE} &= - y \log \sigma \left( S \left(x\right)\right), \\ 
    \mathcal{L} &=  (1 - \lambda_{KL}) \mathcal{L}_{CE} +  \lambda_{KL} \mathcal{L}_{KL}, 
\end{align}
where  $\lambda_{KL}$ is the hyper-parameter controlling the weight of knowledge distillation objective. Recent explorations also find that introducing KD objectives of alignments between the intermediate representations~\citep{romero2014fitnets,Sun2019PatientKD} and attention map~\citep{Jiao2019TinyBERT,wang2020MiniLM} is helpful. 
Note that conventional KD framework is static, i.e., the teacher model is selected before KD and the training is conducted on all training instances indiscriminately according to the pre-defined objective and the corresponding weights of different objectives.
However, it is unreasonable to conduct the KD learning procedure statically as the student model evolves during the training.
We are curious whether adaptive adjusting the settings on teacher adoption, dataset selection and supervision adjustment can bring benefits regarding student performance and learning efficiency, motivating us to explore a dynamic KD framework.

\section{Dynamic Knowledge Distillation}
In this section, we introduce the dynamic knowledge distillation.
The core idea behind is to empower the student to adjust the learning procedure according to its current state, and we investigate the three aspects illustrated in Figure~\ref{fig:dkd}.


\begin{table}[t]
    \centering
    \small 
    \begin{tabular}{@{}l|cccc@{}}
    \toprule 
     \textbf{Method}   & \textbf{RTE} & \textbf{IMDB} & \textbf{CoLA} & \textbf{Avg.}  \\
      \midrule 
     BERT\BASESIZE &  67.8&  89.1
 &  54.2  &  70.4\\ 
     BERT\LARGESIZE  &  72.6 &  90.4& 60.1 &  74.4\\
     \midrule 
    No KD &  63.7 &  86.3& 39.0 & 63.0\\  
    KD w/ BERT\BASESIZE & 64.9 & {86.9}  & 39.4 & 63.7   \\ 
    KD w/ BERT\LARGESIZE & 64.5 & 86.5 & 38.2  &  63.1\\ 
    KD w/ Ensemble & 64.9 & 86.7 & 39.9 & 63.8  \\ 
    \midrule 
    Uncertainty-Hard & \textbf{66.9}$^*$\hspace{-\lsuperstar} & 86.3 & \textbf{42.7}$^*$\hspace{-\lsuperstar} & \textbf{65.3}\\ 
    Uncertainty-Soft& 66.4$^*$\hspace{-\lsuperstar} & \textbf{87.1}$^*$\hspace{-\lsuperstar} & 41.0 &  64.8 \\  
     \bottomrule
    \end{tabular}
    \caption{We find that bigger teacher with better performance raises a worse student model. Results are average of 3 seeds on the validation set. $^*$ denotes statistically significant improvement over the best performing baseline with $p < 0.05$.}
    \label{tab:bigger_not_better}
\end{table}

\subsection{Dynamic Teacher Adoption}
The teacher model plays a vital role in KD, as it provides the student soft-targets for helping the student learn the relation between different classes~\citep{Hinton2015Distilling}.
However, there are few investigations regarding how to select a proper teacher for the student in KD for PLMs during the training dynamically.
In the KD of PLMs, it is usually all teacher models are with the same model architecture, i.e., huge Transformer~\citep{vaswani2017transformer} model. 
Thus, the most informative factor of teacher models is their model size, e.g., the layer number of the teacher model and the corresponding hidden size.
This motivates us to take a first step to explore the impact of model size.

\subsubsection{Bigger Teacher Not Always Raises Better Student}
\label{sec:teacher_size}
Specifically, we are curious about whether learning from a bigger PLM with better performance can lead to a better distilled student model.
We conduct probing experiments to distill a 6-layer student BERT model from BERT\BASESIZE with 12 layers, and BERT\LARGESIZE with 24 layers, respectively.
We conducts the experiment on two datasets, RTE~\citep{bentivogli2009rte} and CoLA~\citep{warstadt-etal-2019cola}, where two teacher models exhibit clear performance gap, and a sentiment classification benchmark IMDB~\citep{IMDB}.
Detailed experimental setup can be found in Appendix~\ref{apx:teacher_size}.

As shown in Table~\ref{tab:bigger_not_better},
we surprisingly find that while the BERT\LARGESIZE teacher clearly outperforms the small BERT\BASESIZE teacher model, the student model distilled by the BERT\BASESIZE teacher achieves better performance on all three datasets. 
This phenomenon is counter-intuitive as a larger teacher is supposed to provide better supervision signal for the student model.
We think that there are two possible factors regarding the size of teacher model that leading to the deteriorated performance:

(1) The predicted logits of the teacher model become less soft as the teacher model becomes larger and more confident about its prediction~\citep{guo2017calibration,desai-durrett-2020calibrationPLM}, which decreases the effect of knowledge transfer via the soft targets. 
We find that a smaller $\tau$ also leads to a decreased performance of the student model, indicating the the less-softened teacher prediction will decrease the student performance.\footnote{Refer to Appendix~\ref{apx:kd_temper} for details.} 

(2) The capacity gap between the teacher and student model increases as the teacher becomes larger. The competency of the student model can not match that of the large teacher model, which weakens the performance of KD. 

\begin{figure}[t!]
    \centering
    \includegraphics[width=0.95\linewidth]{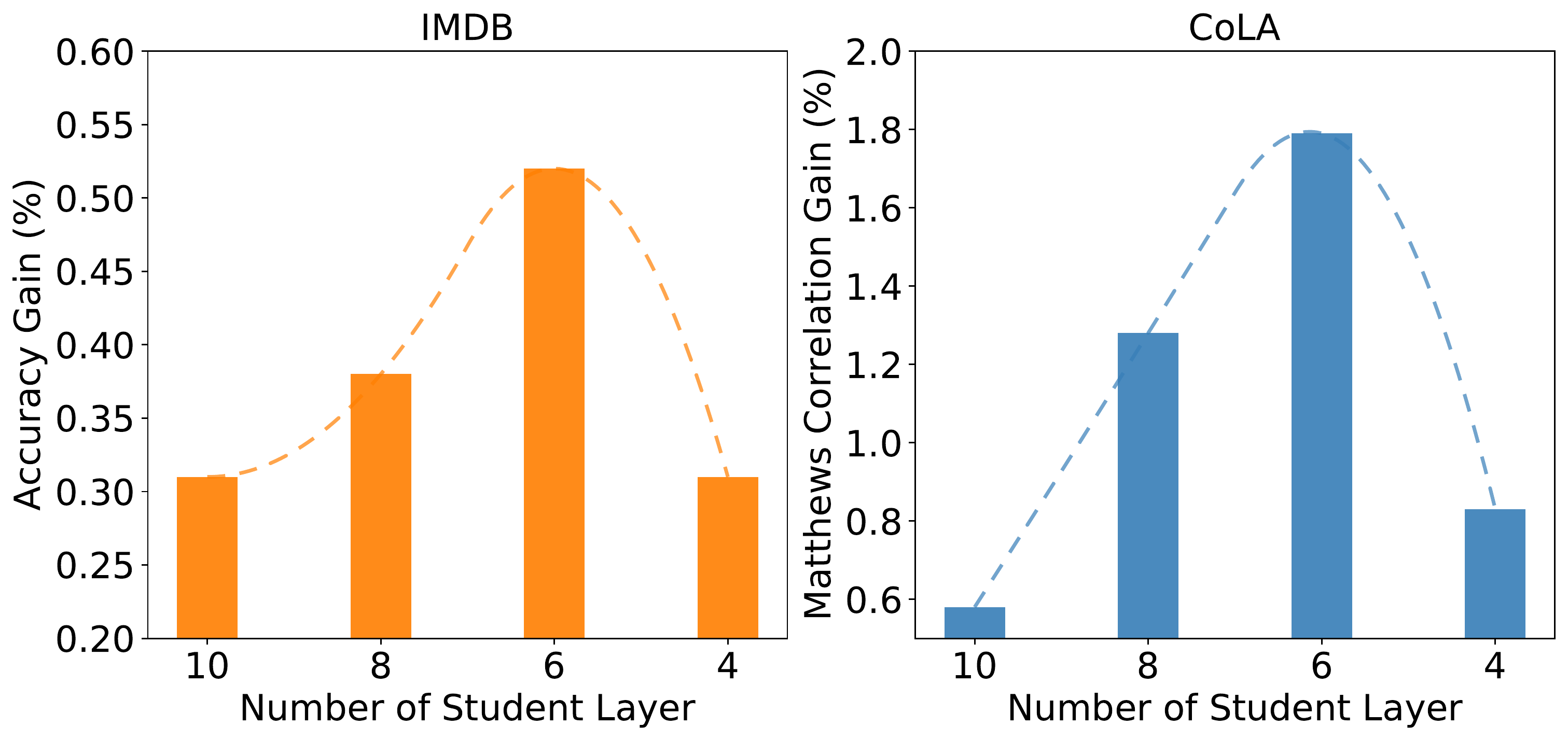}
    \caption{Performance gain of distilled student model with various layer sizes. The teacher model is BERT\BASESIZE with 12 layers.}
    \label{fig:kd_over_ce}
\end{figure}

To explore the combined influence of these factors, we 
distill student models with different layers
and plot the performance gain compared to directly training the student model without distillation in Figure~\ref{fig:kd_over_ce}.
It can be found that by decreasing the student size, the better supervision from teacher model boosts the performance, while the two counteractive factors dominate as the gap becomes much larger, decreasing the performance gain.
We notice that this phenomenon is also observed by ~\citet{Mirzadeh2020TAKD} in computer vision tasks using convolutional networks, showing that it is a widespread issue and needs more in-depth investigations.
Note that BERT\BASESIZE and BERT\LARGESIZE also differs from the number of hidden size, the experiments regarding the hidden size, where the phenomenon also exists and corresponding results can be found in Appendix~\ref{apx:kd_hidden}.

\subsubsection{Uncertainty-based Teacher Adoption}
Our preliminary observations demonstrate that selecting a proper teacher model for KD is significant for the student performance. 
While the capacity gap is an inherent problem once the teacher and the student are set, we are curious about whether dynamically querying the proper teacher according to the student competency during training can make the full use of teacher models.
Without loss of generality, we conduct KD to train a student model from two teacher models with different numbers of Transformer layers.
We assume that during the initial training stage, the student can rely more on the small teacher model, while turns to the large teacher for more accurate supervision when it becomes stronger.
Specifically, we propose to utilize the student prediction uncertainty as a proxy of the competency, inspired by the successful applications in active learning~\citep{settles2009active}, and design two uncertain-based teacher adoption strategies:

\noindent\textbf{Hard Selection}: 
The instances in one batch are sorted according to the student prediction uncertainty, i.e., the entropy of predicted class distribution. 
Then the instances are evenly divided into instances that the student most uncertain about and instances that model is most confident about. 
For the uncertain part, the small teacher is queried for supervision signals, while the large teacher provides the soft-label for the instances that the student is confident about.

\noindent\textbf{Soft Selection}:  The corresponding KD loss weights from two teachers are adjusted softly at instance-level.
Formally, given two teacher model $T_1$~(BERT\BASESIZE, in our case) and $T_2$~(BERT\LARGESIZE), we can re-write the KD objective in the multiple teacher setting as:
\begin{equation}
    \mathcal{L}_{KD} = w_1 \mathcal{L}^{T_1}_{KL} + w_2 \mathcal{L}^{T_2}_{KL}
\end{equation}
where $\mathcal{L}^{T}_{KL}$ denotes matching loss of the output logits of the student model and the teacher model $t$.
The $w_1$ and $w_2$ controls the relative contribution of the supervisions from the two teachers.
We adaptively down-weight the supervision from the large teacher when the student are uncertain about the training instances.
The prediction uncertainty is adopted as a measurement of the student competency for instance $x$:
\begin{equation}
    u_x = \text{Entropy} \left( \sigma \left( S \left(x\right)\right)\right)
\end{equation}
where $\sigma$ is a normalization function, e.g, softmax function for mapping the logit vector to probability distribution.
The $w_1$ and $w_2$ are adjusted as follows:
\begin{align}
    w_1 = \frac{u_x}{U}, \quad  w_2 = 1 - \frac{u_x}{U}
\end{align}
where $U$ is a normalization factor which re-scales the weight to $[0, 1]$.
In this way, the student will pay more attention to the small teacher when it is uncertain about the current instance, while relies on the large teacher when it is confident about its prediction.
\subsubsection{Experiments}
\paragraph{Settings}
We conduct experiments to distill a $6$-layer student model from BERT\BASESIZE and BERT\LARGESIZE, on RTE, CoLA and IMDB, following settings of probing analysis in Section~\ref{sec:teacher_size}.
\paragraph{Results}
The results of the proposed selection strategies are shown in Table~\ref{tab:bigger_not_better}. 
We observe that the hard selection strategy achieves an overall 65.3 accuracy which outperforms directly learning from the ensemble of two teacher models.
This demonstrates that the proposed strategy is effective by selecting the proper teacher model to learn.
The soft selection strategy also outperforms the baseline while falls little behind with the hard version.
We attribute it to that the provided supervisions of two teachers are of different softness, thus may confuse the student model. 

\subsection{Dynamic Data Selection}
The second research question we want to explore is which data will be more beneficial for the performance of the student.
As the distillation proceeds, the student is becoming stronger, thus the repetitive learning on the instances those the student has mastered on can be eliminated.
If there are such instances that are vital for the learning of the student model, can we only conduct KD on these instances, for improving the learning efficiency?
Besides, do the vital instances remain static or also evolve with the student model?
These questions motivate us to explore the effect of dynamically selecting instances.
\subsubsection{Uncertainty-based Data Selection}
We propose to actively select informative instances according to student prediction uncertainty, inspired by the successful practice of active learning~\citep{settles2009active}. 
Formally, given $N$ instances in one batch, for each instance $x$,  the corresponding output class probability distribution over the class label $y$ of the student model is $P(y \mid x ) = \sigma \left( S\left(x\right)\right)$.
We compute an uncertainty score $u_x$ for $x$ using the follow strategies with negligible computational overhead:

\noindent\textbf{Entropy}~\citep{settles2009active}, which measures the uncertainty of the student prediction distribution:
\begin{equation}
    u_x = - \sum_y P\left( y \mid x\right) \log P\left( y \mid x\right).
\end{equation}

\noindent\textbf{Margin}, which is computed as the margin between the first and second most probable class $y_1^*$ and $y_2^*$:
\begin{equation}
    u_x =  P\left(  y_1^* \mid x\right) -  P\left( y_2^* \mid x\right).
\end{equation}

\noindent\textbf{Least-Confidence~(LC)}, which indicates how uncertain the model about the predicted class $\hat y = \arg\max_y P(y \mid x)$:
\begin{align}
    u_x &=  1 - P\left( \hat y \mid x \right) 
\end{align}
We rank the instances in a batch according to its prediction uncertainty, and only choose the top $N \times r$ instances to query the teacher model, where $r \in (0, 1]$ is the selection ratio controlling the number to query. 
Note that in binary classification tasks like IMDB, the selected subsets using the above strategies are identical.
We also design a \textbf{Random} strategy that selects $N \times r$ instances randomly, to serve as a baseline for evaluating the effectiveness of selection strategies.


\begin{table}[t!]
    \centering
    \setlength{\tabcolsep}{2pt}
    \scalebox{0.83}{ 
    \small 
    \begin{tabular}{@{}l|c|rrrr@{}}
    \toprule 
     \textbf{Method} & \textbf{\#FLOPs} & \textbf{SST-5} & \textbf{IMDB} &  \textbf{MRPC} & \textbf{MNLI-m / mm}   \\
      \midrule 
   BERT\BASESIZE~(Teacher)  & - &  52.0 & 89.1  
 &  86.8  &  84.0 / 84.4 \\ 
   Vanilla KD & 45.1B & 47.4  & 86.8 & 80.2  & 81.7 / 82.0 \\  
   \midrule 
   Random &  22.6B& 46.8 & 86.4 &  79.7 & 81.4 / 81.6 \\ 
    Uncertainty-Entropy & 28.2B &46.7 & 86.8 &  79.4&  81.5 / 82.0\\ 
    Uncertainty-Margin & 28.2B  & 46.6 &86.8 &  79.4 &  81.4 / 81.9  \\ 
    Uncertainty-LC & 28.2B & 46.5 & 86.8 & 79.4 &  81.4 / 81.9\\
    \midrule 
    $\Delta$  & - &  - 0.6  & 0.0  & - 0.5 & - 0.2 / 0.0  \\ 
    \bottomrule
    \end{tabular}}
    \caption{Dynamic data selection results with $r$ set to $0.5$. Results are averaged of $3$ seeds on the validation set.
    $\Delta$ denotes the minimal performance degradation of different selection strategies compares to vanilla KD.}
    \label{tab:selection_normal_dataset}
\end{table}

\begin{table}[t!]
    \centering
    \scalebox{0.8}{
    \begin{tabular}{@{}lrrrrc@{}}
    \toprule
        \textbf{Dataset} & \textbf{\# Train}  &  \textbf{\# Aug Train }&\textbf{\# Dev} & \textbf{\# Test} &   \textbf{\# Class}  \\
        \midrule
                 SST-5 & 8.8k & 176k  & 1.1k & 2.2k& 5   \\ 
               IMDB &  20k   & 400k & 5k& 25k& 2  \\ 
         MNLI &  393k   & 786,0k &20k &20k  &  3   \\ 
         MRPC &   3.7k & 74k & 0.4k & 1.7k &  2  \\ 
          RTE &   2.5k  &  50k &0.3k &3k  &  2  \\ 
          CoLA &  8.5k & 170k & 1k &1k  & 2 \\ 
         \bottomrule
    \end{tabular}}
    \caption{Statistics of datasets. \textbf{\# Aug Train} denotes the number of the augmented training dataset following \citet{Jiao2019TinyBERT}.
    }
    \label{tab:dataset}
\end{table}

\begin{table*}[!t]
    \centering
    \small 
    \begin{tabular}{@{}l |  c  | c c c c c c  @{}}
    \toprule
     \textbf{Method}  &  \textbf{\#FLOPs}  & \textbf{SST-5} &  \textbf{IMDB}  & \textbf{MRPC} & \textbf{MNLI-m / mm}&  \textbf{Avg.} ($\uparrow$) & \textbf{$\Delta$} ($\downarrow$)\\
     \midrule
    BERT\BASESIZE~(Teacher) & -    &  53.7 & 88.8   &  87.5& 83.9 / 83.4  & 79.5  &  -\\
    \midrule 
    $\text{TinyBERT}^\dagger$ &   24.9B  &  -&  - &  86.4 &  82.5 / 81.8 & - &  - \\
    TinyBERT &  24.9B    & 51.4 & 87.6 & 86.2 & 82.6 / 82.0 &  78.0   & 0.0 \\
    \midrule 
    Random &  2.49B  &51.1  & 87.0  &  83.3 &  80.8 / 80.5   &  76.5  &  1.5\\
    Uncertainty-Entropy & 4.65B  &51.5  &\textbf{87.7}  &   \textbf{86.5}&  \textbf{81.8} / 81.0   & \textbf{77.7} & \textbf{0.3} \\ 
    Uncertainty-Margin& 4.65B  & \textbf{51.6}  & \textbf{87.7}  & \textbf{86.5} & 81.6 / \textbf{81.1}&  \textbf{77.7}& \textbf{0.3}\\ 
    Uncertainty-LC  & 4.65B   & 51.2 & \textbf{87.7}  & \textbf{86.5}&  81.4 / 80.8 & 77.5  & 0.5\\
    \bottomrule
    \end{tabular}
    \caption{Test results when the selection ratio $r=0.1$ for dynamic data selection on various tasks. \#FLOPs denotes the average computational cost of KD for each instance.
    $\dagger$ denotes results from~\citet{Jiao2019TinyBERT}.}
    \label{tab:main_ret}
\end{table*}

\subsubsection{Experiments}
\paragraph{Settings}
We conduct the investigation experiments on two sentiment classification datasets IMDB~\citep{IMDB} and SST-5~\citep{socher2013sst}, and natural language inference tasks including MRPC~\citep{dolan2005mrpc} and MNLI~\citep{williams2018mnli}. 
The statistics of dataset and the implementation details can be found in Table~\ref{tab:dataset} and \ref{apx:implementation}, respectively.
We report accuracy as the performance measurement for all the evaluated tasks.

Besides, we also provide the corresponding computational FLOPs for comparing the learning efficiency. In more detail, we divide the computational cost $C$ of KD tinto three parts: student forward $F_s$, teacher forward $F_t$ and student backward $B_s$ for updating parameters. 
Note that $F_s \approx B_s  \ll F_t$, as the teacher model is usually much larger than the student model. 
By actively learning only from $N \times r$ instances that the student are most uncertain about, the cost is reduced to:
\begin{equation}
    C' = N \times F_s + N \times r \times B_s + N \times r \times F_t. 
\end{equation}
For example, the number of computational FLOPs of a 6-layer student BERT model is $11.3$B while that of a 12-layer BERT teacher model is $22.5$B~\citep{Jiao2019TinyBERT}. 
When $r$ is set to $0.1$, the total KD cost is reduced from $45.1$B to $14.7$B.


\paragraph{Results with Original Dataset}
The results when $r$ set to $0.5$ are listed in Table~\ref{tab:selection_normal_dataset}.
Overall, it can be found that selecting the instances only lead negligible degradation of the student performance, compared to that of Vanilla KD, showing the effectiveness of the uncertainty-based strategies. Interestingly, the random strategy perform closely to the uncertainty-based strategies, which we attribute to that the underlying informative data space can also be covered by random selected instances.
Besides, we notice that performance drop is smaller on the tasks with larger training data size.
For example, selecting informative instances with prediction entropy leads to 0.2 accuracy drop on the MNLI dataset consisting $393$k training instances, while causes 0.8 performance drop on MRPC with $3.7$k instances.
A possible reason is that for the tiny dataset, the underlying data distribution is not well covered by the training data, therefore further down-sampling the training data results in a larger performance gap.
To verify this, we turn to the the setting where the original training dataset is enriched with augmentation techniques.
\paragraph{Results with Augmented Dataset}
Following TinyBERT~\citep{Jiao2019TinyBERT}, we augment the training dataset 20 times with BERT mask language prediction, as it has been prove effective for distilling a powerful student model.
Our assumption is that with the data augmentation technique, the training set can sufficiently cover the possible data space, thus selecting the informative instances will not lead to significant performance drop.
Besides, it is of great practical value to accelerate the KD procedure via reducing the queries to teacher model on the augmented dataset.
For example, it costs about $\$3,709$ for querying all the instances of the augmented MNLI dataset as mentioned by \citet{krishna2019thieves}.\footnote{The cost is estimated according to Google Cloud natural language API: \url{https://cloud.google.com/natural-language/pricing}.}
By only querying a small portion of instances to the teacher model, we can greatly reduce economic cost and ease the possible environmental side-effects~\citep{strubell2019energy,schwartz2019greenai,xu21knas} that may hinder the deployments of PLMs on downstream tasks.

\begin{figure}
    \centering
    \includegraphics[width=0.9\linewidth]{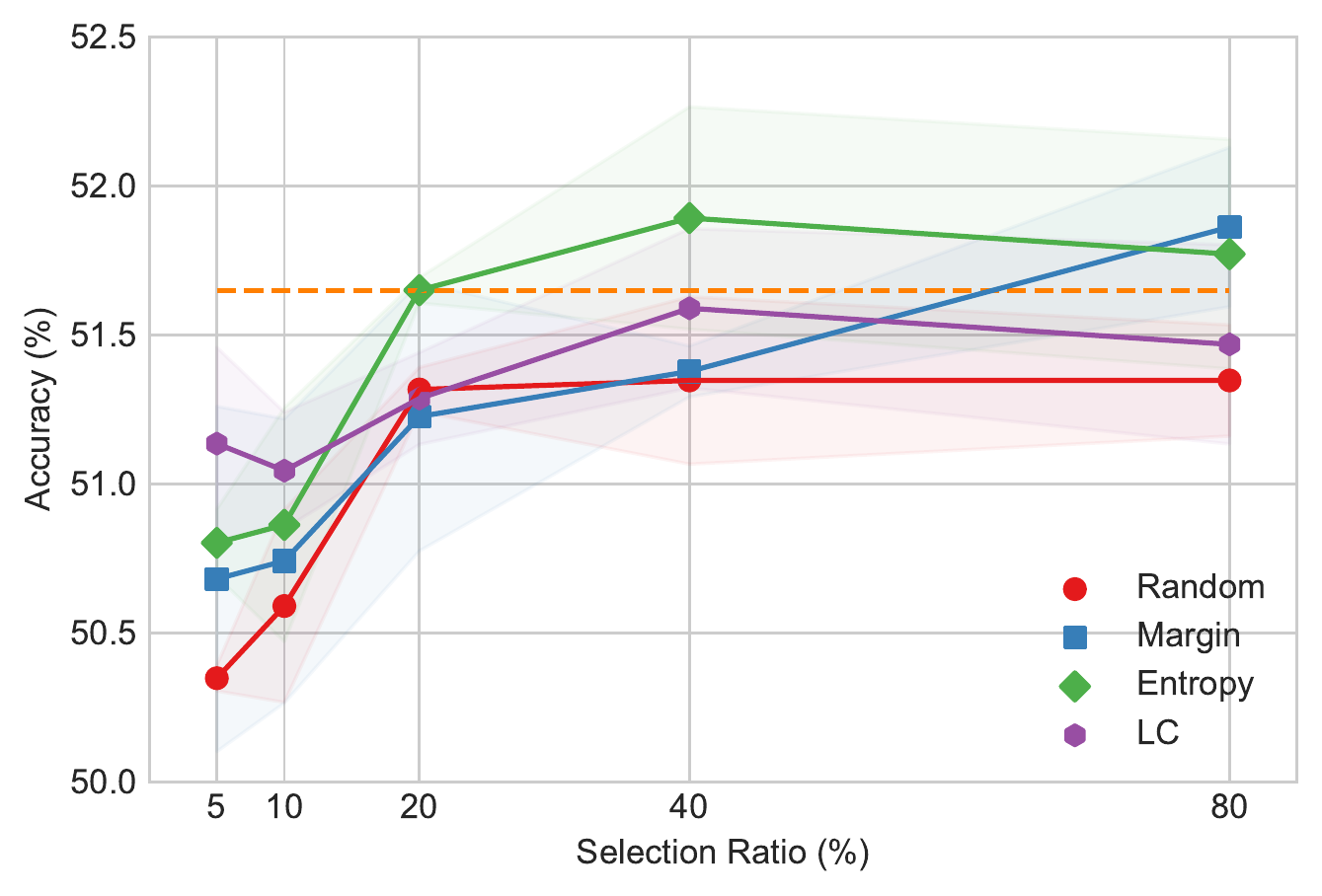}
    \caption{We plot the mean accuracy on the validation set of 3 seeds~($\pm$ one standard deviation) under different selection ratios of various strategies. Orange dashed line denotes the performance of vanilla KD.}
    \label{fig:varying_ratio}
\end{figure}

The results with TinyBERT-4L as the backbone model and $r=0.1$ are listed in Table~\ref{tab:main_ret}.
We can observe that uncertainty-based selection strategy can maintain the superior performance while saving the computational cost, e.g., the FLOPs is reduced from $24.9$B to $4.65$B with negligible average performance decrease. 
In tasks like SST-5 and IMDB, selecting $10\%$ most informative instances according to student prediction entropy can even outperform the original TinyBERT using the whole dataset.
Among these strategies, the least-confidence strategy performs relatively poor, as it only takes the maximum probability into consideration while neglects the full output distribution.
\paragraph{Performance under Different Ratios}
We vary the selection ratio $r$ to check the results of different strategies on the augmented SST-5 dataset.
The results are shown in Figure~\ref{fig:varying_ratio}.
Our observations are:
(1) There exists a trade-off between the performance and the training costs, i.e., increasing the selection ratio generally improves the performance of student model, while results in bigger training costs.
(2) We can achieve the full performance using about 20\% training data. It indicates that the training data support can be well covered with about $20$\% data, thus learning from these instances can sufficiently train the student model.
It validates our motivation to select informative instances for reducing the repetitive learning caused by data redundancy.
(3) Selection strategies based on the uncertainty of student prediction can make the better use of limited query, performing better than the random selection, especially when the query number is low.

\subsubsection{Analysis}
\begin{figure}[t!]
    \centering
    \includegraphics[width=0.98\linewidth]{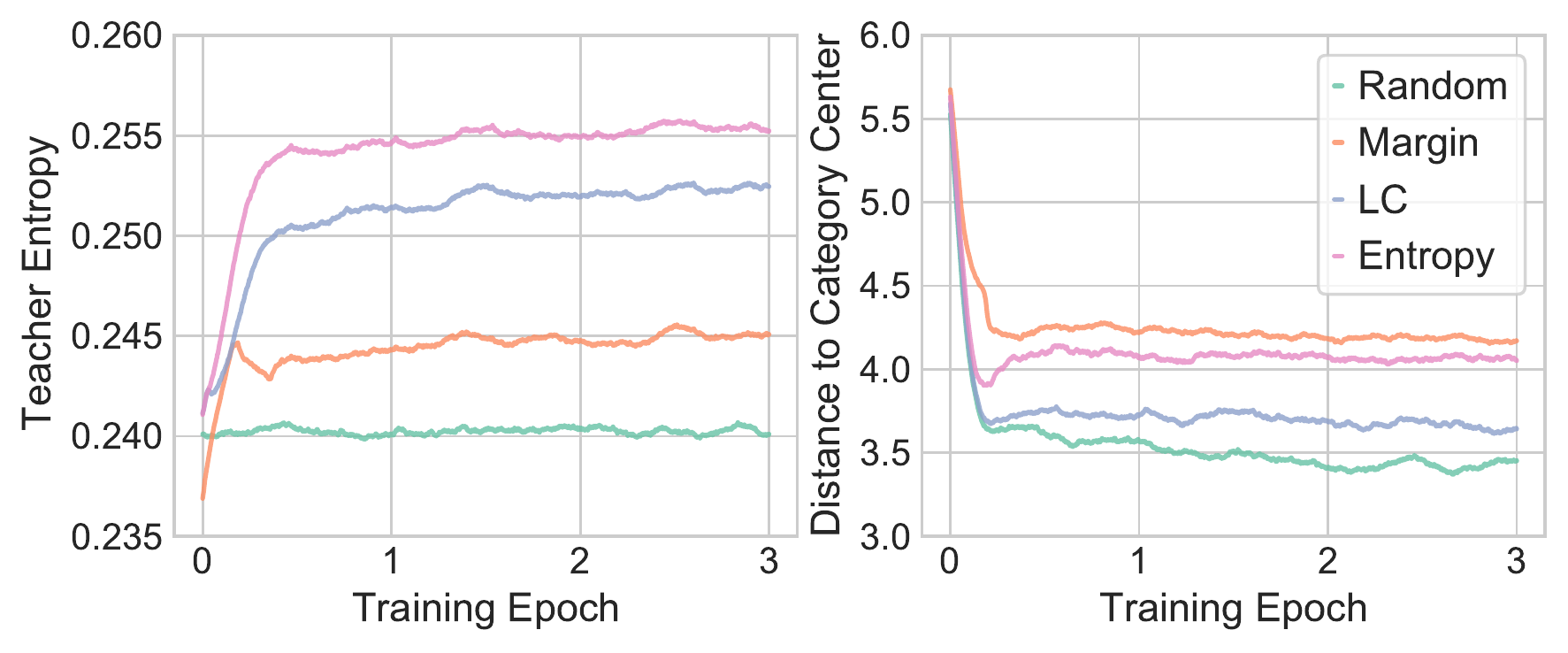}
    \caption{Training dynamics comparison of the selected instances. The entropy strategy can distinguish more informative instances from the whole dataset.}
    \label{fig:entropy_curve}
\end{figure}

We further conduct experiments on the augmented SST-5 dataset to gain insights about the property of selected instances and visualize the distribution of selected instances for intuitive understanding.
\paragraph{Property of Selected Instances}
We plot the teacher prediction entropy and the distance from the selected instances to the corresponding category center.
From Figure~\ref{fig:entropy_curve}~(left), we observe
for hard instances with high uncertainty that selected by the student model, the teacher model also regards them as difficult. It indicates that the instance difficulty is an inherent property of data and uncertainty-based criterion can discover these hard instances from the whole dataset.
Besides, the teacher's entropy of selected instances increases as the training proceeds, showing that the selected instances also evolve during the training as the student is becoming stronger. 
The right part in Figure~\ref{fig:entropy_curve} demonstrates that uncertainty-based selection will pick up the instances that are far away from the category center than the ones are randomly picked.
These instances are more informative for the student model to learn the decision boundary of different classes.

\paragraph{Visualization of Selected Instances}

\begin{figure}[t]
  \centering
  \subfigure[Random]{\includegraphics[width=0.49\linewidth]{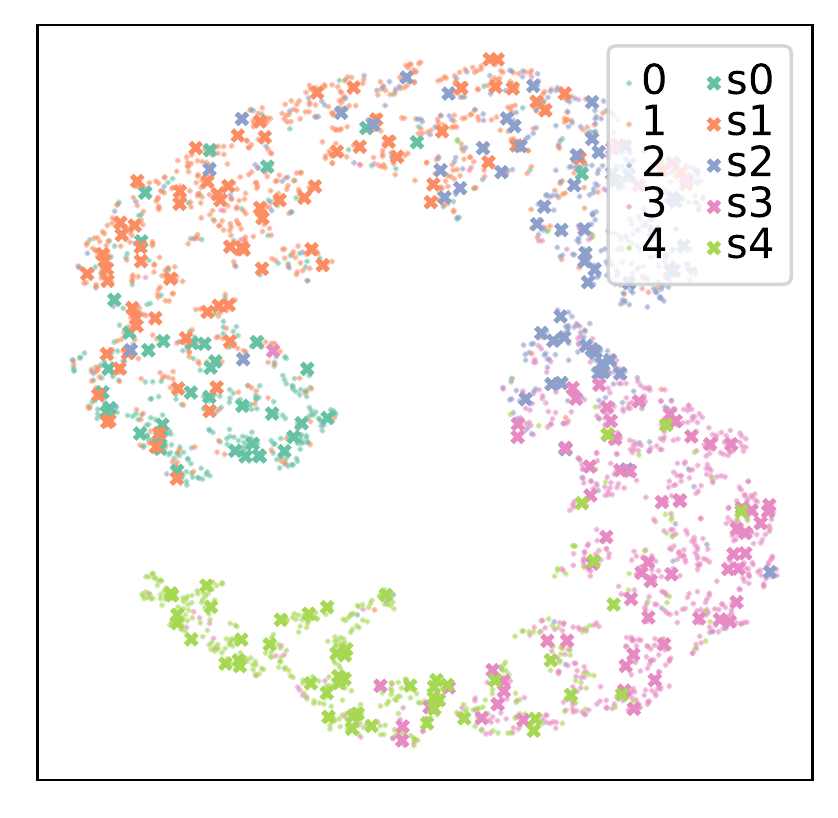}}
    \subfigure[Margin]{\includegraphics[width=0.49\linewidth]{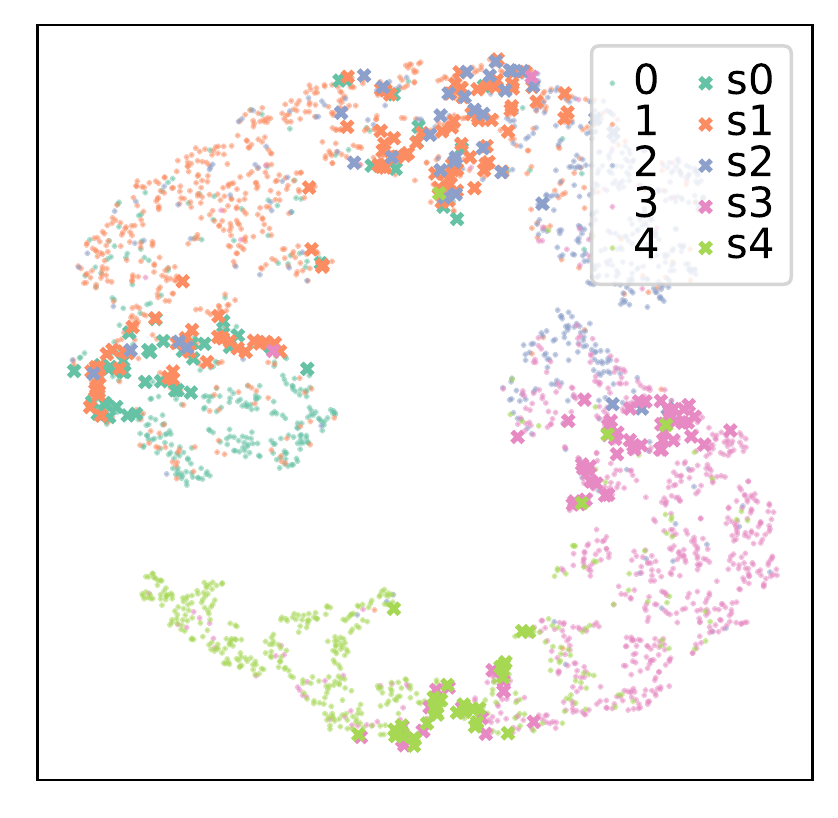}}
  \caption{The t-SNE visualization of instance representations. Uncertainty-based strategies select the instances close to the class boundary, which is useful for the learning of the student model.
  Best viewed in color.
  }
  \label{fig:umap_vis}
\end{figure}
We further visualize the distribution of instances in the feature space, i.e., the representation before the classifier layer, using t-SNE~\citep{maaten2008tsne} and highlight the selected instances with the cross marker.
We compare the best performing strategy margin and random selection on SST-5.
As shown in Figure~\ref{fig:umap_vis}, the instances randomly selected are distributed uniformly in feature space. 
The margin strategy instead picks the instances close to the classification boundary.
The results demonstrate that the uncertainty-based selection criterion can help the student model pay more attention to the instances that are vital for making correct predictions, thus achieving a comparable performance with a much lower computational cost.

In all, our analysis experiments show that the uncertainty-based selection is effective for picking instances that are close to the classification boundary. Besides, the selected instances also evolve as the student model becomes stronger.
By learning from these instances, the computational cost of KD is greatly reduced with a negligible accuracy drop.

\subsection{Dynamic Supervision Adjustment}
We finally explore the question of the optimal learning objective functions.
Previous studies have shown that integrating the alignments on the intermediate representation~\citep{romero2014fitnets,Sanh2019DistilBERT,Sun2019PatientKD} and attention map~\citep{Jiao2019TinyBERT,wang2020MiniLM} between the student and the teacher model can further boost the performance.
We are interested in whether the dynamic adjustment of the supervision from different alignment objectives can bring extra benefits.
As the first exploration, we only consider the combination of the KL-divergence distance with the teacher prediction and the hidden representation alignments:
\begin{equation}
    \mathcal{L}_{KD} = \lambda_{KL} * \mathcal{L}_{KL} + \lambda_{PT} * \mathcal{L}_{PT}
\end{equation}
where $\mathcal{L}_{PT}$ is called PaTient loss, which measures the alignment between the normalized internal representations of the teacher and student model~\citep{Sun2019PatientKD}:
\begin{equation}
\mathcal{L}_{PT}=\sum_{i=1}^{M}\left\|\frac{\mathbf{h}_{i}^{s}}{\left\|\mathbf{h}_{i}^{s}\right\|_{2}}-\frac{\mathbf{h}_{I_{p t}(j)}^{t}}{\left\|\mathbf{h}_{ I_{p t}(j)}^{t}\right\|_{2}}\right\|_{2}^{2}
\end{equation}
where $M$ is the number of student layer, $I_{pt}(i)$ denotes the corresponding alignment of the teacher layer for the student $i$-th layer, $\mathbf{h}_i^s$ and $\mathbf{h}_i^t$ denote representation of $i$-th layer of student and teacher model, respectively.

\subsubsection{Uncertainty-based Supervision Adjustment}
Different from previous studies which set the corresponding alignment objective weights via hyper-parameter search and keep them unchanged during the training, we propose to adjust the weights according to the student prediction uncertainty for each instance.
The motivation behind is that we assume it is unnecessary to force the student model to align all the outputs of the teacher model during the whole training stage.
As the training proceeds, the student is become stronger and it may learn the informative features different from the teacher model.
Therefore, there is no need to force the student to act exactly with the teacher model, i.e., requiring the intermediate representations of the student to be identical with the teacher.
Formally, we turn the weight of KD objective into a function of the student prediction uncertainty $u(x) = \text{Entropy} \left ( \sigma\left ( S \left( s\right)\right) \right) $:
\begin{align}
    \lambda_{KL} &= \lambda^*_{KL}  ( 1-  \frac{u_x}{U} ), \quad \lambda_{PT} = \lambda^{*}_{PT} \frac{u_x}{U} 
\end{align}
where $\lambda^{*}_{KL}$ and $\lambda^{*}_{PT}$ are pre-defined weight for each objective obtained by parameter search and $U$ is the normalization factor. 
In this way, the contribution of these two objectives are adjusted dynamically during the training for each instances.
For instances that the student is confident about, the supervision from the internal representation alignment is down-weighted. Thus the student is focusing mimicking the final prediction probability distribution with the teacher based on its own understanding of the instance.
On the contrary, for instances that the student is confusing, the supervision from teacher model representations can help it learn the feature of the instance better.

\subsubsection{Experiments}
\paragraph{Settings}
The student model is set to 6-layer and BERT\BASESIZE is adopted as the teacher model.
For intermediate layer representation alignment, we adopt the Skip strategy, i.e., $I_{pt} = \{2, 4, 6, 8, 10\}$ as it performs best as described in BERT-PKD.
We conduct experiments on the sentiment analysis task SST-5, and two natural language inference tasks MRPC and RTE. For $\lambda^{*}_{KL}$ and $\lambda^{*}_{PT}$, we adopt the searched parameters provided by~\citet{Sun2019PatientKD}.

\begin{table}[t!]
    \centering
    \small 
    \begin{tabular}{@{}l|cccc@{}}
    \toprule
     \textbf{Method}  &  \textbf{SST-5}& \textbf{MRPC}& \textbf{RTE} & \textbf{Avg.}\\
     \midrule
     BERT\BASESIZE~(Teacher) & 52.0 & 86.8   & 67.8 & 68.9  \\ 
     \midrule 
     Vanilla KD & 47.4& 80.2   &  64.9 & 64.2 \\ 
     BERT-PKD & 46.6 & 80.8 & 65.1 &  64.2 \\ 
     Uncertainty & \textbf{48.1}& \textbf{81.5}$^*$\hspace{-\lsuperstar} & \textbf{66.4}$^*$\hspace{-\lsuperstar} & \textbf{65.3}\\ 
    \bottomrule
    \end{tabular}
    \caption{Results of dynamic adjusting the supervision weights, showing the uncertainty-based adjustment is effective. $^*$ denote results are statistically significant with $p < 0.05$.}
    \label{tab:dynamic_supervision}
\end{table}

\begin{figure}
    \centering
    \includegraphics[width=0.98\linewidth]{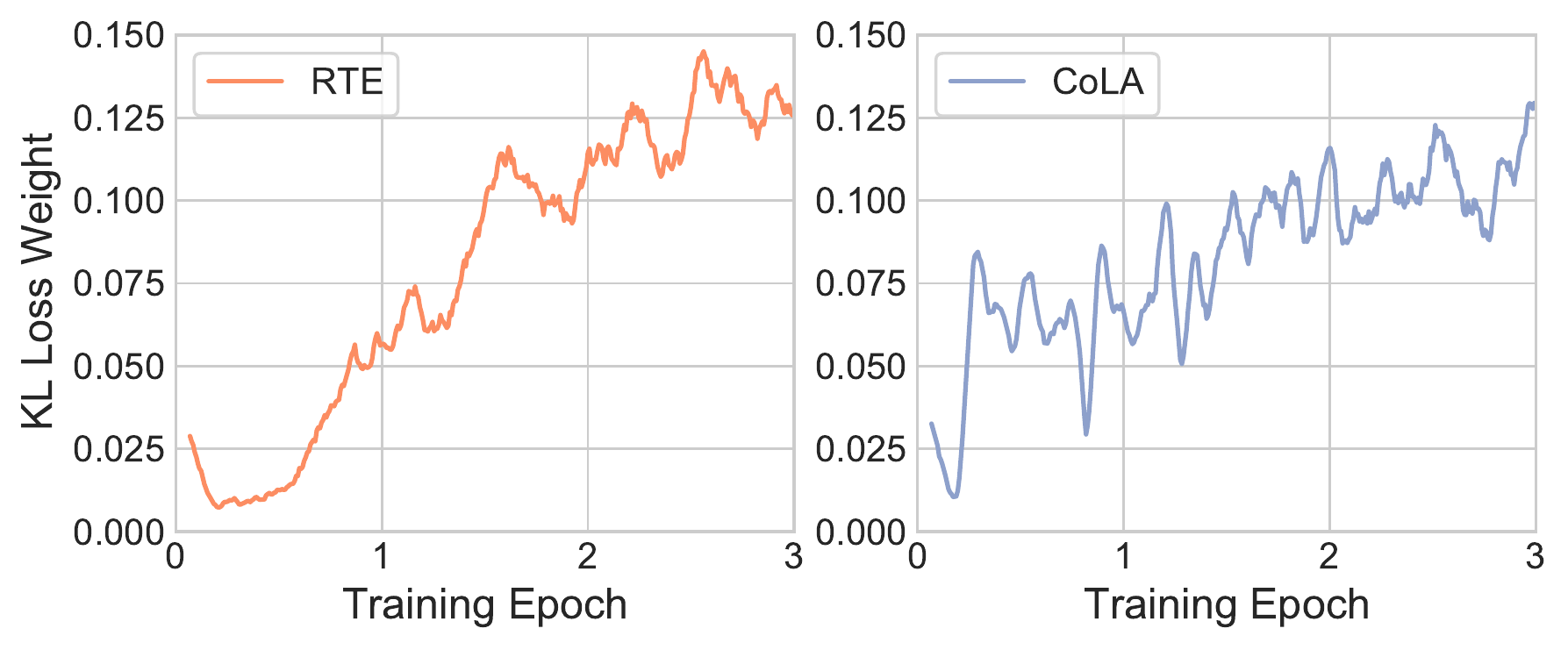}
    \caption{Evolution of the dynamically adjusted weight of KL-divergence loss weight.}
    \label{fig:ds_curve}
\end{figure}

\paragraph{Results}
The results of adaptive adjusting the supervision weights are listed in Table~\ref{tab:dynamic_supervision}.
We observe that the proposed uncertainty-based supervision adjustment can outperform the static version BERT-PKD on all the tasks, showing that the proper adjustment of the KD objectives is effective for improving the student performance.
We also plot the batch average of the KL loss weight in Figure~\ref{fig:ds_curve}.
As expected, the corresponding weight of the prediction probability alignment objective is increasing as the student becomes more confident about its predictions, thus paying more attention to matching the output distribution with the teacher model.
Interestingly, we find that at the initial stage of training, the KL weight is decreasing. It indicates that the learning by aligning the intermediate representations can help the student quickly gain the understanding the task, thus improving the confidence of predictions.


\section{Discussions}
After the preliminary explorations on the three aspects of Dynamic KD, we observe that it is promising for improving the efficiency and the distilled student performance.
Here we provide potential directions for further investigations.

(1) From uncertainty-based selection criterion to advanced methods. In this paper, we utilize student prediction uncertainty as a proxy for selecting teachers, training instances and supervision objectives. More advanced methods based on more accurate uncertainty estimations~\citep{gal2016dropout,zhou2020uncertainty}, clues from training dynamics~\citep{toneva2018empirical}, or even a learnable selector can be developed.

(2) From isolation to integration. As a preliminary study, we only investigate the three dimensions independently. Future work can adjust these components simultaneously and investigate the underlying correlation between these three dimensions for a better efficiency-performance trade-off.

(3) More fine-grained investigations regarding different components in the Dynamic KD framework: (i) For teacher adoption, exploring whether dynamically training a student model from more teacher models or teacher models with different architectures can bring extra benefits; (ii) For data selection, it will be interesting to investigate whether the informative data is model-agnostic, and whether dynamically selecting data from different domains can improve the generalization performance; (iii) For supervision adjustment, investigations on the effect of combinations of different objectives can be promising.
\section{Related Work}
Our work relates to recent explorations on applying KD for compressing the PLMs and active learning.
\paragraph{Knowledge Distillation for PLMs}
Knowledge distillation~\citep{Hinton2015Distilling} aims to transfer the dark knowledge from a large teacher model to a compact student model, which has been proved effective for obtaining compact variants of PLMs.
Those methods can be divided into general distillation~\citep{Sanh2019DistilBERT,turc2019well,wang2020MiniLM} and task-specific distillation~\citep{Sun2019PatientKD,Jiao2019TinyBERT,Xu2020BERTofTheseus,li2020bertemd,liang2020mixkd,li2020accelerating,wu-etal-2021-one}.
The former conducts KD on the general text corpus while the latter trains the student model on the task-specific datasets. 
In this paper, we focus on the latter one as it is more widely adopted in practice. Compared to existing static KD work, we are the first to explore the idea of Dynamic KD, making it more flexible, efficient and effective.




\paragraph{Active Learning}~\citep{settles2009active}, where a learning system is allowed to choose the data from which it learns from, for achieving better performance with fewer labeled data. 
Traditional selection strategies include uncertainty-based methods~\citep{scheffer2001activeMargin,settles2009active}, which select the instances that model is most uncertain about, query-by-committee~\citep{freund1997qbc}, which select instances with highest disagreements between a set of classifiers, and methods based on decision theory~\citep{roy2001error}. In this paper, inspired by the success of active learning, we introduce Dynamic KD that utilizes the different strategies like prediction entropy as a proxy of student competency to adaptively adjust the different aspects of KD. Our explorations show that the uncertainty-based strategies are effective for improving the efficiency and performance of KD.

\section{Conclusion}
In this paper, we introduce dynamic knowledge distillation, and conduct exploratory experiments regarding teacher model adoption, data selection and the supervision adjustment.
Our experimental results demonstrate that the dynamical adjustments on the three aspects according to the student uncertainty is promising for improving the student performance and learning efficiency.
We provide discussions on the potential directions worth exploring in the future, and hope this work can motivate studies towards more environmental-friendly knowledge distillation methods.

\section*{Acknowledgements}
We thank all the anonymous reviewers for their constructive comments and Xuancheng Ren for his valuable suggestions in preparing the manuscript.
This work was supported by a Tencent Research Grant. 
Xu Sun is the corresponding author of this paper.

\nocite{wolf-etal-2020-transformers,loshchilov2018adamw}

\bibliography{anthology,custom}

\begin{thebibliography}{38}
\expandafter\ifx\csname natexlab\endcsname\relax\def\natexlab#1{#1}\fi

\bibitem[{Bentivogli et~al.(2009)Bentivogli, Dagan, Hoa, Giampiccolo, and
  Magnini}]{bentivogli2009rte}
Luisa Bentivogli, Ido~Kalman Dagan, Dang Hoa, Danilo Giampiccolo, and Bernardo
  Magnini. 2009.
\newblock The fifth pascal recognizing textual entailment challenge.
\newblock In \emph{TAC Workshop}.

\bibitem[{Desai and Durrett(2020)}]{desai-durrett-2020calibrationPLM}
Shrey Desai and Greg Durrett. 2020.
\newblock \href {https://www.aclweb.org/anthology/2020.emnlp-main.21}
  {Calibration of pre-trained transformers}.
\newblock In \emph{EMNLP}, pages 295--302, Online.

\bibitem[{Devlin et~al.(2019)Devlin, Chang, Lee, and
  Toutanova}]{devlin2019bert}
Jacob Devlin, Ming{-}Wei Chang, Kenton Lee, and Kristina Toutanova. 2019.
\newblock \href {https://www.aclweb.org/anthology/N19-1423} {{BERT}:
  Pre-training of deep bidirectional transformers for language understanding}.
\newblock In \emph{NAACL-HLT}, pages 4171--4186.

\bibitem[{Dolan and Brockett(2005)}]{dolan2005mrpc}
William~B. Dolan and Chris Brockett. 2005.
\newblock \href {https://www.aclweb.org/anthology/I05-5002} {Automatically
  constructing a corpus of sentential paraphrases}.
\newblock In \emph{Proceedings of the Third International Workshop on
  Paraphrasing ({IWP})}.

\bibitem[{Freund et~al.(1997)Freund, Seung, Shamir, and Tishby}]{freund1997qbc}
Yoav Freund, H~Sebastian Seung, Eli Shamir, and Naftali Tishby. 1997.
\newblock Selective sampling using the query by committee algorithm.
\newblock \emph{Machine learning}, 28(2):133--168.

\bibitem[{Gal and Ghahramani(2016)}]{gal2016dropout}
Yarin Gal and Zoubin Ghahramani. 2016.
\newblock Dropout as a bayesian approximation: Representing model uncertainty
  in deep learning.
\newblock In \emph{ICML}, pages 1050--1059.

\bibitem[{Guo et~al.(2017)Guo, Pleiss, Sun, and
  Weinberger}]{guo2017calibration}
Chuan Guo, Geoff Pleiss, Yu~Sun, and Kilian~Q Weinberger. 2017.
\newblock On calibration of modern neural networks.
\newblock In \emph{ICML}, pages 1321--1330.

\bibitem[{Hinton et~al.(2015)Hinton, Vinyals, and Dean}]{Hinton2015Distilling}
Geoffrey Hinton, Oriol Vinyals, and Jeff Dean. 2015.
\newblock Distilling the knowledge in a neural network.
\newblock \emph{arXiv preprint arXiv:1503.02531}.

\bibitem[{Jiao et~al.(2020)Jiao, Yin, Shang, Jiang, Chen, Li, Wang, and
  Liu}]{Jiao2019TinyBERT}
Xiaoqi Jiao, Yichun Yin, Lifeng Shang, Xin Jiang, Xiao Chen, Linlin Li, Fang
  Wang, and Qun Liu. 2020.
\newblock \href {https://www.aclweb.org/anthology/2020.findings-emnlp.372}
  {{T}iny{BERT}: Distilling {BERT} for natural language understanding}.
\newblock In \emph{Findings of EMNLP}, pages 4163--4174.

\bibitem[{Krishna et~al.(2019)Krishna, Tomar, Parikh, Papernot, and
  Iyyer}]{krishna2019thieves}
Kalpesh Krishna, Gaurav~Singh Tomar, Ankur~P Parikh, Nicolas Papernot, and
  Mohit Iyyer. 2019.
\newblock Thieves on sesame street! {M}odel extraction of {BERT}-based {API}s.
\newblock In \emph{ICLR}.

\bibitem[{Li et~al.(2020{\natexlab{a}})Li, Liu, Zhao, Xu, Yang, and
  Jin}]{li2020bertemd}
Jianquan Li, Xiaokang Liu, Honghong Zhao, Ruifeng Xu, Min Yang, and Yaohong
  Jin. 2020{\natexlab{a}}.
\newblock \href {https://www.aclweb.org/anthology/2020.emnlp-main.242}
  {{BERT}-{EMD}: Many-to-many layer mapping for {BERT} compression with earth
  mover{'}s distance}.
\newblock In \emph{EMNLP}, pages 3009--3018.

\bibitem[{Li et~al.(2020{\natexlab{b}})Li, Lin, Ren, Chen, Ren, Li, Zhou, and
  Sun}]{li2020accelerating}
Lei Li, Yankai Lin, Shuhuai Ren, Deli Chen, Xuancheng Ren, Peng Li, Jie Zhou,
  and Xu~Sun. 2020{\natexlab{b}}.
\newblock Accelerating pre-trained language models via calibrated cascade.
\newblock \emph{arXiv preprint arXiv:2012.14682}.

\bibitem[{Liang et~al.(2021)Liang, Hao, Shen, Zhou, Chen, Chen, and
  Carin}]{liang2020mixkd}
Kevin~J. Liang, Weituo Hao, Dinghan Shen, Yufan Zhou, Weizhu Chen, Changyou
  Chen, and Lawrence Carin. 2021.
\newblock {MixKD}: Towards efficient distillation of large-scale language
  models.
\newblock In \emph{ICLR}.

\bibitem[{Liu et~al.(2019)Liu, Ott, Goyal, Du, Joshi, Chen, Levy, Lewis,
  Zettlemoyer, and Stoyanov}]{Liu2019RoBERTa}
Yinhan Liu, Myle Ott, Naman Goyal, Jingfei Du, Mandar Joshi, Danqi Chen, Omer
  Levy, Mike Lewis, Luke Zettlemoyer, and Veselin Stoyanov. 2019.
\newblock Ro{BERT}a: {A} robustly optimized {BERT} pretraining approach.
\newblock \emph{arXiv preprint arXiv:1907.11692}.

\bibitem[{Loshchilov and Hutter(2018)}]{loshchilov2018adamw}
Ilya Loshchilov and Frank Hutter. 2018.
\newblock Decoupled weight decay regularization.
\newblock In \emph{{ICLR}}.

\bibitem[{Maas et~al.(2011)Maas, Daly, Pham, Huang, Ng, and Potts}]{IMDB}
Andrew~L. Maas, Raymond~E. Daly, Peter~T. Pham, Dan Huang, Andrew~Y. Ng, and
  Christopher Potts. 2011.
\newblock \href {https://www.aclweb.org/anthology/P11-1015/} {Learning word
  vectors for sentiment analysis}.
\newblock In \emph{ACL}, pages 142--150.

\bibitem[{Maaten and Hinton(2008)}]{maaten2008tsne}
Laurens van~der Maaten and Geoffrey Hinton. 2008.
\newblock Visualizing data using {t-SNE}.
\newblock \emph{JMLR}, 9:2579--2605.

\bibitem[{Mirzadeh et~al.(2020)Mirzadeh, Farajtabar, Li, Levine, Matsukawa, and
  Ghasemzadeh}]{Mirzadeh2020TAKD}
Seyed{-}Iman Mirzadeh, Mehrdad Farajtabar, Ang Li, Nir Levine, Akihiro
  Matsukawa, and Hassan Ghasemzadeh. 2020.
\newblock Improved knowledge distillation via teacher assistant.
\newblock In \emph{AAAI}, pages 5191--5198.

\bibitem[{Romero et~al.(2015)Romero, Ballas, Kahou, Chassang, Gatta, and
  Bengio}]{romero2014fitnets}
Adriana Romero, Nicolas Ballas, Samira~Ebrahimi Kahou, Antoine Chassang, Carlo
  Gatta, and Yoshua Bengio. 2015.
\newblock {FitNets}: Hints for thin deep nets.
\newblock In \emph{{ICLR}}.

\bibitem[{Roy and McCallum(2001)}]{roy2001error}
Nicholas Roy and Andrew McCallum. 2001.
\newblock Toward optimal active learning through sampling estimation of error
  reduction.
\newblock In \emph{ICML}, pages 441--448.

\bibitem[{Sanh et~al.(2019)Sanh, Debut, Chaumond, and
  Wolf}]{Sanh2019DistilBERT}
Victor Sanh, Lysandre Debut, Julien Chaumond, and Thomas Wolf. 2019.
\newblock Distil{BERT}, a distilled version of {BERT}: smaller, faster, cheaper
  and lighter.
\newblock In \emph{NeurIPS Workshop on Energy Efficient Machine Learning and
  Cognitive Computing}.

\bibitem[{Scheffer et~al.(2001)Scheffer, Decomain, and
  Wrobel}]{scheffer2001activeMargin}
Tobias Scheffer, Christian Decomain, and Stefan Wrobel. 2001.
\newblock Active hidden markov models for information extraction.
\newblock In \emph{International Symposium on Intelligent Data Analysis}, pages
  309--318. Springer.

\bibitem[{Schwartz et~al.(2019)Schwartz, Dodge, Smith, and
  Etzioni}]{schwartz2019greenai}
Roy Schwartz, Jesse Dodge, Noah~A Smith, and Oren Etzioni. 2019.
\newblock Green {AI}.
\newblock \emph{arXiv preprint arXiv:1907.10597}.

\bibitem[{Settles(2009)}]{settles2009active}
Burr Settles. 2009.
\newblock Active learning literature survey.
\newblock Computer Sciences Technical Report 1648, University of
  Wisconsin--Madison.

\bibitem[{Socher et~al.(2013)Socher, Perelygin, Wu, Chuang, Manning, Ng, and
  Potts}]{socher2013sst}
Richard Socher, Alex Perelygin, Jean Wu, Jason Chuang, Christopher~D Manning,
  Andrew~Y Ng, and Christopher Potts. 2013.
\newblock \href {https://www.aclweb.org/anthology/D13-1170} {Recursive deep
  models for semantic compositionality over a sentiment treebank}.
\newblock In \emph{EMNLP}, pages 1631--1642.

\bibitem[{Strubell et~al.(2019)Strubell, Ganesh, and
  McCallum}]{strubell2019energy}
Emma Strubell, Ananya Ganesh, and Andrew McCallum. 2019.
\newblock \href {https://www.aclweb.org/anthology/P19-1355} {Energy and policy
  considerations for deep learning in {NLP}}.
\newblock In \emph{ACL}, pages 3645--3650.

\bibitem[{Sun et~al.(2019)Sun, Cheng, Gan, and Liu}]{Sun2019PatientKD}
Siqi Sun, Yu~Cheng, Zhe Gan, and Jingjing Liu. 2019.
\newblock \href {https://www.aclweb.org/anthology/D19-1441} {Patient knowledge
  distillation for {BERT} model compression}.
\newblock In \emph{EMNLP-IJCNLP}, pages 4323--4332.

\bibitem[{Toneva et~al.(2018)Toneva, Sordoni, des Combes, Trischler, Bengio,
  and Gordon}]{toneva2018empirical}
Mariya Toneva, Alessandro Sordoni, Remi~Tachet des Combes, Adam Trischler,
  Yoshua Bengio, and Geoffrey~J Gordon. 2018.
\newblock An empirical study of example forgetting during deep neural network
  learning.
\newblock In \emph{ICLR}.

\bibitem[{Turc et~al.(2019)Turc, Chang, Lee, and Toutanova}]{turc2019well}
Iulia Turc, Ming-Wei Chang, Kenton Lee, and Kristina Toutanova. 2019.
\newblock Well-read students learn better: The impact of student initialization
  on knowledge distillation.
\newblock \emph{arXiv preprint arXiv:1908.08962}.

\bibitem[{Vaswani et~al.(2017)Vaswani, Shazeer, Parmar, Uszkoreit, Jones,
  Gomez, Kaiser, and Polosukhin}]{vaswani2017transformer}
Ashish Vaswani, Noam Shazeer, Niki Parmar, Jakob Uszkoreit, Llion Jones,
  Aidan~N. Gomez, Lukasz Kaiser, and Illia Polosukhin. 2017.
\newblock Attention is all you need.
\newblock In \emph{NeurIPS}, pages 5998--6008.

\bibitem[{Wang et~al.(2020)Wang, Wei, Dong, Bao, Yang, and
  Zhou}]{wang2020MiniLM}
Wenhui Wang, Furu Wei, Li~Dong, Hangbo Bao, Nan Yang, and Ming Zhou. 2020.
\newblock {MiniLM}: Deep self-attention distillation for task-agnostic
  compression of pre-trained transformers.
\newblock In \emph{NeurIPS}.

\bibitem[{Warstadt et~al.(2019)Warstadt, Singh, and
  Bowman}]{warstadt-etal-2019cola}
Alex Warstadt, Amanpreet Singh, and Samuel~R. Bowman. 2019.
\newblock \href {https://aclanthology.org/Q19-1040} {Neural network
  acceptability judgments}.
\newblock \emph{TACL}, 7:625--641.

\bibitem[{Williams et~al.(2018)Williams, Nangia, and Bowman}]{williams2018mnli}
Adina Williams, Nikita Nangia, and Samuel Bowman. 2018.
\newblock \href {https://www.aclweb.org/anthology/N18-1101} {A broad-coverage
  challenge corpus for sentence understanding through inference}.
\newblock In \emph{NAACL-HLT}, pages 1112--1122.

\bibitem[{Wolf et~al.(2020)Wolf, Debut, Sanh, Chaumond, Delangue, Moi, Cistac,
  Rault, Louf, Funtowicz, Davison, Shleifer, von Platen, Ma, Jernite, Plu, Xu,
  Scao, Gugger, Drame, Lhoest, and Rush}]{wolf-etal-2020-transformers}
Thomas Wolf, Lysandre Debut, Victor Sanh, Julien Chaumond, Clement Delangue,
  Anthony Moi, Pierric Cistac, Tim Rault, R{\'{e}}mi Louf, Morgan Funtowicz,
  Joe Davison, Sam Shleifer, Patrick von Platen, Clara Ma, Yacine Jernite,
  Julien Plu, Canwen Xu, Teven~Le Scao, Sylvain Gugger, Mariama Drame, Quentin
  Lhoest, and Alexander~M. Rush. 2020.
\newblock \href {https://www.aclweb.org/anthology/2020.emnlp-demos.6}
  {Transformers: State-of-the-art natural language processing}.
\newblock In \emph{System Demonstrations, {EMNLP}}, pages 38--45.

\bibitem[{Wu et~al.(2021)Wu, Wu, and Huang}]{wu-etal-2021-one}
Chuhan Wu, Fangzhao Wu, and Yongfeng Huang. 2021.
\newblock One teacher is enough? pre-trained language model distillation from
  multiple teachers.
\newblock In \emph{Findings of ACL-IJCNLP 2021}, pages 4408--4413.

\bibitem[{Xu et~al.(2020)Xu, Zhou, Ge, Wei, and Zhou}]{Xu2020BERTofTheseus}
Canwen Xu, Wangchunshu Zhou, Tao Ge, Furu Wei, and Ming Zhou. 2020.
\newblock \href {https://www.aclweb.org/anthology/2020.emnlp-main.633}
  {{BERT}-of-{Theseus}: Compressing {BERT} by progressive module replacing}.
\newblock In \emph{EMNLP}, pages 7859--7869.

\bibitem[{Xu et~al.(2021)Xu, Zhao, Lin, Gao, Sun, and Yang}]{xu21knas}
Jingjing Xu, Liang Zhao, Junyang Lin, Rundong Gao, Xu~Sun, and Hongxia Yang.
  2021.
\newblock {KNAS:} green neural architecture search.
\newblock In \emph{ICML}, pages 11613--11625.

\bibitem[{Zhou et~al.(2020)Zhou, Yang, Wong, Wan, and
  Chao}]{zhou2020uncertainty}
Yikai Zhou, Baosong Yang, Derek~F Wong, Yu~Wan, and Lidia~S Chao. 2020.
\newblock \href {https://aclanthology.org/2020.acl-main.620} {Uncertainty-aware
  curriculum learning for neural machine translation}.
\newblock In \emph{ACL}, pages 6934--6944.

\end{thebibliography}
\bibliographystyle{acl_natbib}

\appendix
\section{Teacher Size Exploration Settings}
\label{apx:teacher_size}
We conduct the knowledge distillation with BERT\BASESIZE and BERT\LARGESIZE as teacher models. The student model is set to a 6-layer student BERT.
For training the teacher, the teacher model is fine-tuned using the script provided by Huggingface Transformers library. The fine-tuning learning rate is $2\text{e-}5$ with a linear warm-up learning rate schedule for the first $10\%$ training steps. Batch size is $32$, training epoch is set to $3$, and the max length of input sentence is set to $128$. The statistics of used datasets are listed in Table~\ref{tab:dataset}. 

For distilling the student model, the student model is initialized using the first $6$ layers weights of BERT\BASESIZE. We adopt the KL-divergence distance as the KD objective.
$\lambda_{KL}$ is set to $0.5$ and we empirically find this setting works well.
The same training hyper-parameters as fine-tuning the teacher model are used for distillation.
The performance is evaluated on the validation set and averaged on $3$ random seeds.

\section{Impacts of Prediction Smoothness}
\label{apx:kd_temper}
\begin{figure}[t]
    \centering
   \subfigure[RTE]{\includegraphics[width=0.45\linewidth]{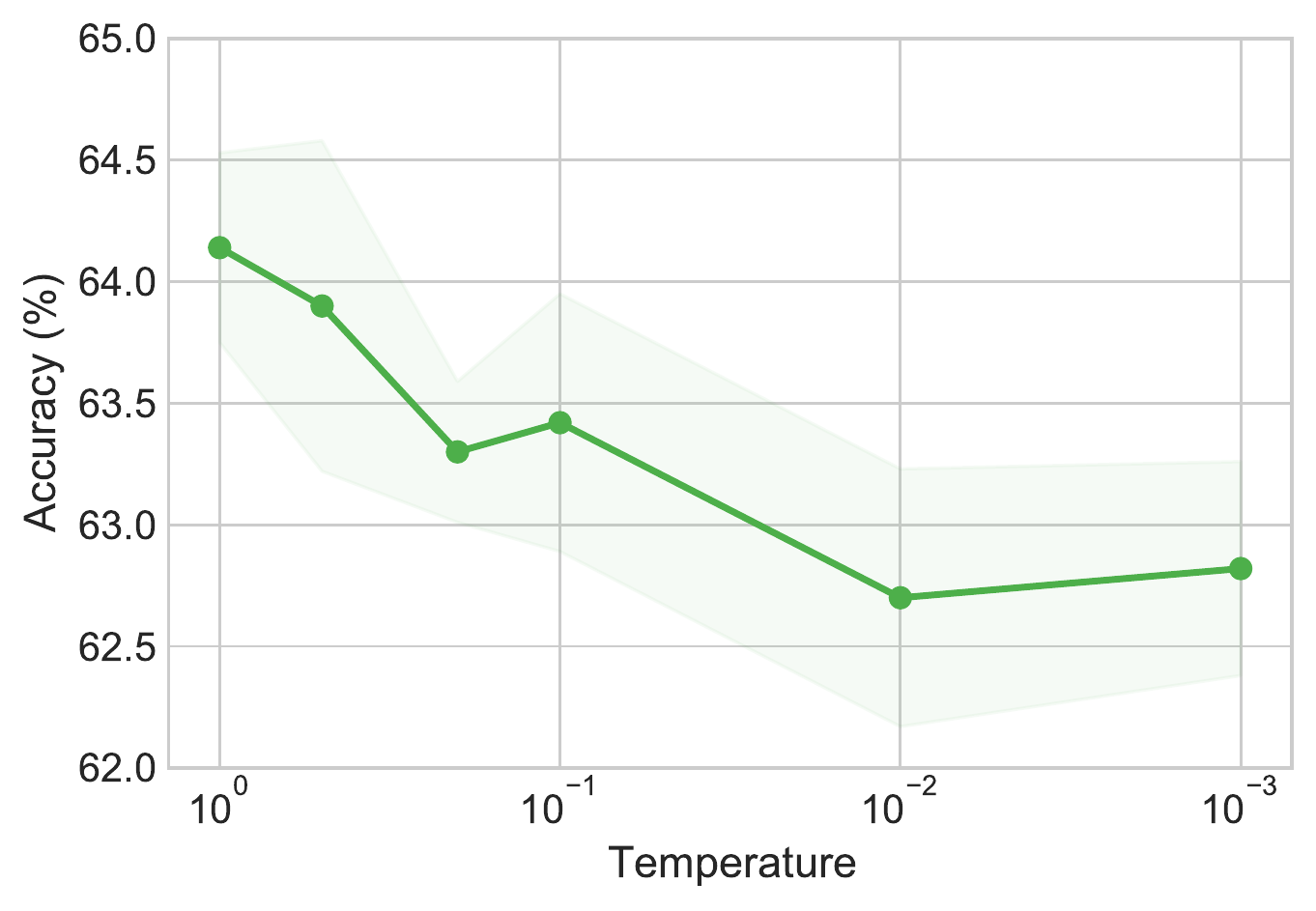}}
   \subfigure[IMDB]{\includegraphics[width=0.45\linewidth]{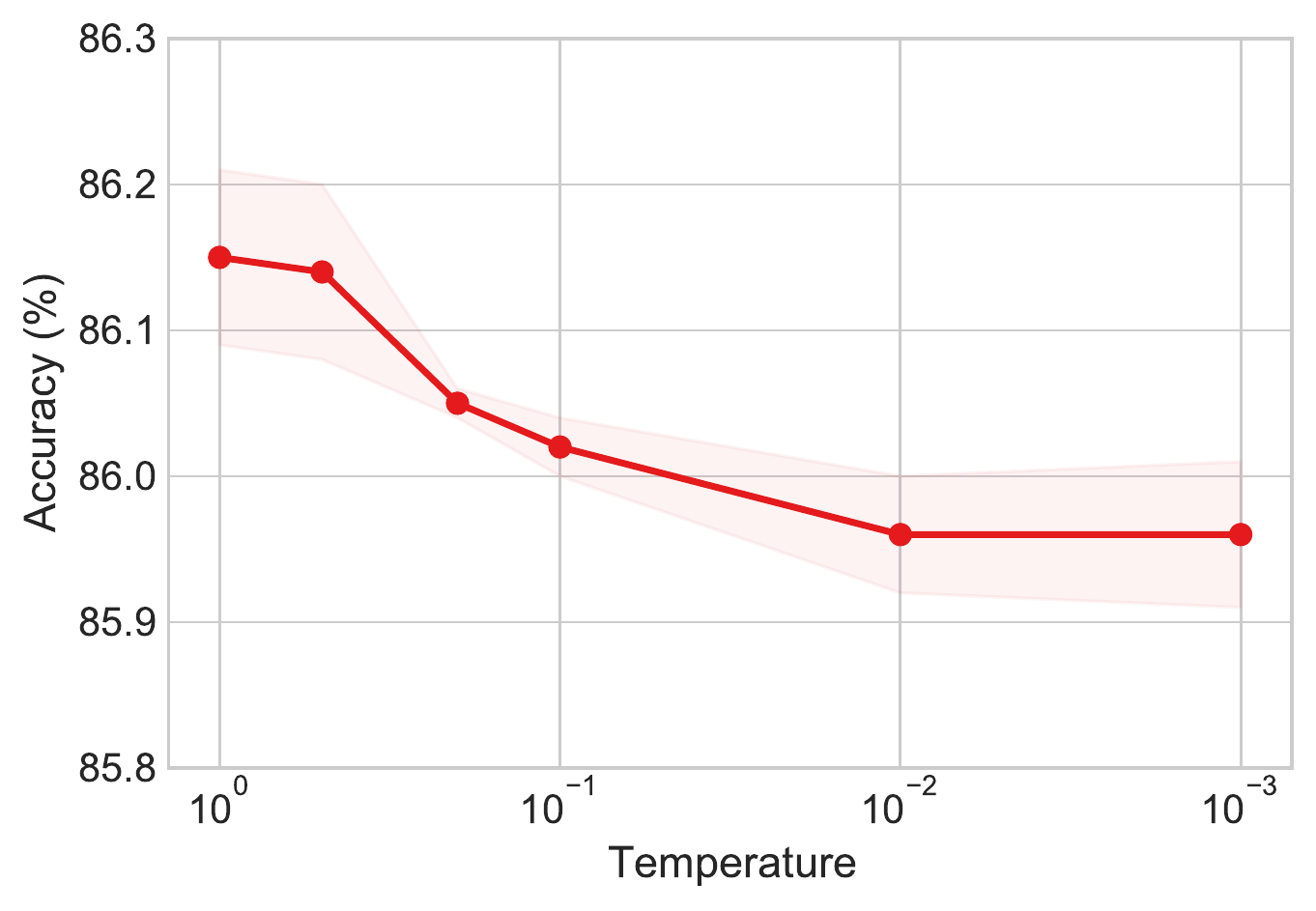}}
    \caption{Distillation performance with varying temperature $\tau$ on different datasets.}
    \label{fig:vary_t}
\end{figure}

To examine the influence of less-softened teacher predictions, we conduct distillation experiments with various temperature $\tau$ using the hyper-parameters identical with previous experiments, to mimicking the sharpen impact introduced by the larger teacher size.
In more detail, we setup a student model with $6$ layers as before and select the BERT\BASESIZE as the teacher model.
The results are illustrated in Figure~\ref{fig:vary_t}.
We observe on both datasets, the decreased temperature $\tau$ will lead a performance decrease.
It indicates that the less-softened probability distribution indeed weakens the performance of knowledge distillation.

\section{Impacts of Teacher Hidden Size}
\label{apx:kd_hidden}
As mentioned in the main paper, we observe that larger teacher may not raise a student model with better performance.
We further conduct experiments regarding the hidden size of teacher model.
Specifically, we setup a student with $6$ layers with $256$ hidden units.
The small teacher and the large teacher are a BERT model of 12-layer with $256$ hidden units and a BERT of 12-layer with $768$ hidden units, respectively.
Out experiments show that on the CoLA dataset, the student model distilled with the small teacher can achieve $11.9$ matthews correlation score while that of model distilled by the large teacher is $8.8$.
The result on the IMDB is consistent, i.e., $83.2$ accuracy for student model distilled by the large teacher and $83.4$ accuracy for the student distilled by the small teacher.
These results again verify the phenomenon that the larger teacher may not always raise a better student model.
     

\section{Implementation Details}
\label{apx:implementation}
Our implementation is based on PyTorch and Huggingface transformers library.
Model is optimized with AdamW optimizer with linear learning rate warm-up. The sentence length is set to $64$ for SST-5 and $128$ for the rest datasets.
Our teacher model is BERT\BASESIZE.
The model is trained with learning rate $2\text{e-}5$ and batch size $32$ for $3$ epochs.
$\lambda_{KL}$ is set to $0.5$, with temperature $\tau$ set to $1$.

For experiments using TinyBERT, we select $\text{TinyBERT}_4$ v2 as our backbone model, and conduct the general distillation for $10$ epochs on the augmented dataset.
We further train the model for $3$ epochs on the augmented dataset and choose learning rates from $\{1\text{e-}5, 2\text{e-}5, 3\text{e-}5\}$ and batch sizes from $\{16, 32\}$ based on the performance on the validation set. The performance are evaluated on the test sets and $\tau$ is set to $1$, following the practice of TinyBERT.

\end{document}